\documentclass[11pt,letterpaper]{article}
\usepackage[left=0.95in,right=0.95in,top=0.83in,bottom=0.85in,headheight=14pt]{geometry}
\usepackage[T1]{fontenc}
\usepackage{mathptmx,microtype,courier}
\usepackage{graphicx,booktabs,tabularx,array,multirow}
\usepackage[table]{xcolor}
\usepackage{amsmath,amssymb,mathtools}
\usepackage[numbers,sort&compress]{natbib}
\usepackage{caption,subcaption,titlesec,enumitem,placeins}
\usepackage{float}
\floatstyle{ruled}
\newfloat{algorithm}{tbp}{loa}
\floatname{algorithm}{Algorithm}
\usepackage[hidelinks]{hyperref}
\newcommand{\doi}[1]{{\urlstyle{same}\href{https://doi.org/#1}{doi: \nolinkurl{#1}}}}
\definecolor{navy}{HTML}{1D3548}
\definecolor{teal}{HTML}{187D83}
\definecolor{pale}{HTML}{F1F5F6}
\definecolor{graytext}{HTML}{52616B}
\titleformat{\section}{\large\bfseries}{\thesection}{0.65em}{}
\titleformat{\subsection}{\normalsize\bfseries}{\thesubsection}{0.6em}{}
\titlespacing*{\section}{0pt}{2.2ex plus .5ex}{0.8ex}
\titlespacing*{\subsection}{0pt}{1.6ex plus .4ex}{0.6ex}
\setlist[itemize]{leftmargin=*,itemsep=2pt,topsep=4pt}
\graphicspath{{figures/}}

\newcolumntype{Y}{>{\raggedright\arraybackslash}X}

\hypersetup{pdftitle={ReuseMem: One Memory, Multiple Goals in Robot Manipulation},pdfauthor={Yoonseok Kang, Dayeon Jung, Seunghoon Jeon, Wonjib Kim, Dongchul Park},pdfsubject={Common-experience memory for robot manipulation}}
\begin{document}
\title{\vspace{-1.8em}\fontsize{21}{24}\selectfont\bfseries ReuseMem: One Memory, Multiple Goals\\in Robot Manipulation\vspace{-0.7em}}
\author{%
\normalsize Yoonseok Kang, Dayeon Jung, Seunghoon Jeon, Wonjib Kim, Dongchul Park\thanks{Corresponding author: \texttt{dongchul@cau.ac.kr}.}\\[0.25em]
\small Chung-Ang University}
\date{}
\maketitle\vspace{-2.5em}
\begin{abstract}
Robots that work alongside people should be able to use what they have seen when new requests arise. The same experience may later support identifying an object or returning it to its original location, although these uses were unknown during observation. ReuseMem learns to use one visual memory for multiple later manipulation goals. It records visual features independently of future goals, retrieves complementary evidence for each request, and combines it with the current scene through a lightweight action interface. Retrieval is supervised by how well candidate evidence supports expert-action prediction across different current states, linking memory selection to its use in manipulation. Repeated visual features are stored once, and memory computations that remain unchanged during action generation are reused. Experiments on four simulation tasks and two physical scenarios demonstrate reuse of a common recording across different requests. With matched action adaptation, ReuseMem improves joint goal completion on the benchmark tasks by 24.5 percentage points over uniform memory. These findings show that one memory can support multiple later manipulation goals as the scene changes.
 Project Page: {\urlstyle{same}\url{https://yoondol.github.io/reusemem/}}.
\end{abstract}
\section{Introduction}
\label{sec:introduction}
Memory enables robots to act on information that is no longer visible. In everyday assistance, the same experience may become useful for several requests that arise only later. An earlier observation may identify an object, reveal its original location, or establish the order of events. The robot need not know which of these uses will matter when it observes the scene. This calls for one memory that supports multiple later goals as the environment changes.

\begin{figure}[H]\centering
\includegraphics[width=\textwidth]{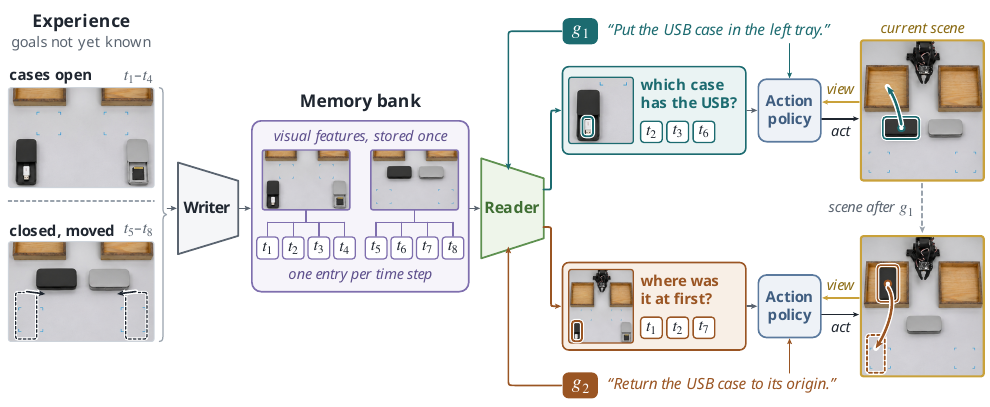}
\caption{\textbf{Overview of ReuseMem.} The robot records case contents and origins before later requests are known. The same memory supplies different historical evidence for each goal, while current observations guide the next manipulation.}
\label{fig:teaser}\end{figure}

Consider a robot observing the contents and original positions of two cases before they are closed and relocated (Figure~\ref{fig:teaser}). Bringing the case containing a USB drive to a tray requires remembering its contents; returning it requires its origin. The same experience supplies different facts for different actions. After the transfer, the case must be grasped from the tray, while its destination still comes from the recorded past.

Memory-augmented policies have advanced historical retrieval and its use in control~\citep{memoryvla,halo,imprint,mem4act,hamlet}. Persistent experience systems also reuse past observations for later requests through retrieval and planning~\citep{memer,memora,messymem,mem}. Building on these advances, we ask how the same recorded experience can reliably support different later manipulations. Success on several tasks with separate histories does not directly answer this question. The relevant unit is a common experience, whose usefulness must persist across requests that need different evidence and produce different actions.

Such reuse couples two challenges. The robot must select the past observations needed for each request, then use them with the current scene to generate an appropriate action. Remembering an object's identity or former position does not specify where to grasp it now. Earlier actions can change the scene while leaving the historical facts valid. Training must therefore connect evidence selection to action under these changing current conditions.

ReuseMem addresses this problem with a goal-independent Writer, a learned Reader, and a lightweight action interface. The Writer retains visual features and separate temporal entries during observation and execution, without an advance inventory of objects or relations. The Reader selects complementary observations for each request. Its supervision comes from comparing how well candidate evidence supports expert-action prediction across different current states for the same goal. Historical facts remain fixed while the required actions vary. The action interface then learns from the updated Reader's selections, combining remembered evidence with current observations through fusion and low-rank adaptation of a frozen pretrained policy.

Practical reuse also requires controlling storage and response time. Recordings often contain many similar views, yet their order can distinguish events relevant to later requests. To limit storage, ReuseMem stores matching visual features once while retaining a separate temporal entry for each observation. To reduce response time, fixed memory projections are computed once per action-sampling call and reused as the action is refined, while the policy still reads the selected memory anew at every step.

Four simulation tasks and two physical scenarios evaluate different goals from common recordings, both independently and in continuous execution. With matched action adaptation, ReuseMem improves joint goal completion by 24.5 percentage points over uniform memory. Ablations and memory interventions examine evidence selection and action use, while storage and response measurements assess the cost of reuse.

Our contributions are:
\begin{itemize}
\item \textbf{One memory, multiple manipulation goals.} We introduce ReuseMem, a framework that records experience before later goals are known and reuses the same visual memory for different manipulations as the current scene changes.
\item \textbf{Action-guided memory learning.} We train memory retrieval by comparing how different selections of past observations support expert-action prediction across current states for the same goal. A lightweight action interface learns to combine the selected evidence with current observations.
\item \textbf{Efficient memory reuse.} We reduce memory storage and repeated computation while retaining the temporal structure of experience. Repeated visual features are shared across temporal entries, and fixed memory projections are reused within each action-sampling call.
\end{itemize}

\section{Related Work}
\label{sec:related}
\paragraph{Memory for robot control.}
Memory augments generalist visuomotor policies~\citep{diffusionpolicy,octo,openvla,pi0} through visual or event-based banks~\citep{memoryvla,dnr,eventvla,sam2act}, recurrent states or test-time adaptation~\citep{membodied,rememvla,muvla,robottt}. Moment tokens, past-action prediction and action-history denoising provide temporal training signals~\citep{hamlet,ptp,actmemvla}, while ECoMEM builds instruction-conditioned concept records~\citep{ecomem}. HALO guides retrieval with video question answering and sparse attention~\citep{halo}. Imprint pretrains multi-stream memory with reconstruction and progress estimation before policy adaptation; its recurrent variant updates memory independently of language~\citep{imprint}. Mem4Act combines semantic and perceptual supervision with memory-first training~\citep{mem4act}. ReuseMem learns preferences between explicit evidence selections from expert-action prediction across current scenes, then trains the action interface with the Reader's deployment selections. MemoryBench, RoboMME and RMBench assess historical dependence~\citep{sam2act,robomme,rmbench}; our evaluation pairs different requests with the same recording.

\paragraph{Experience for later requests.}
Persistent memory supports subsequent reasoning, planning and manipulation~\citep{memer,memora,messymem,mem,pal}. Scene graphs organize observed objects, locations and interactions~\citep{conceptgraphs,roboexp,dynamem,scenegraphil}. MemER reuses keyframes across successive requests through a high-level VLM that issues subgoals to a low-level policy, while MessyMem links interaction-updated scene graphs to visual records for VLM planning~\citep{memer,messymem}. MEMORA edits typed records, MEM combines video and text, and PAL stores semantic events and executable subtasks~\citep{memora,mem,pal}. These systems establish the value of experience beyond its original task. ReuseMem focuses on the learned path from retrieved visual evidence to current action generation, using a common recording to evaluate different manipulation purposes.

\paragraph{Efficient memory and inference.}
Compression and caching address different costs. Token Merging reduces redundancy within visual transformers~\citep{tome}; NativeMEM compresses historical frame views, and MemBodied maintains fixed-size recurrent state~\citep{nativemem,membodied}. MemoryVLA merges similar bank entries~\citep{memoryvla}, while SimpleMemVLA reuses native-video prefixes during inference~\citep{simplememvla}. ReuseMem separates shared appearance storage from temporal occurrences, preserving the order and identity of recorded entries. Its cache covers fixed memory projections within one action-sampling call, while new observations refresh the action input.

\section{Method}\label{sec:method}
\begin{figure}[H]\centering\includegraphics[width=0.98\textwidth]{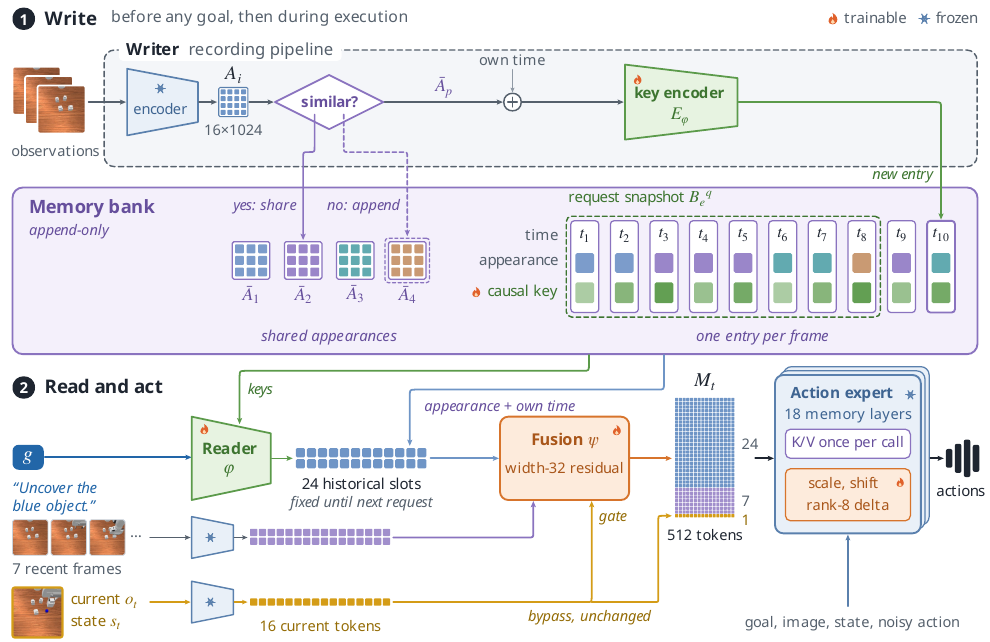}
\caption{\textbf{ReuseMem architecture.} The Writer avoids duplicate appearance storage while retaining a separate temporal entry for each observation. Execution observations pass through the same Writer. Each new request reads the accumulated bank, and current-scene fusion connects selected evidence to the action policy.}\label{fig:architecture}
\end{figure}

ReuseMem records experience before its later goals are known, retrieves evidence for each request, and uses that evidence to act from the current scene (Figure~\ref{fig:architecture}). The Writer preserves visual and temporal information throughout observation and execution. The Reader and action interface learn how to use it.

\paragraph{Problem setting.}\label{sec:problem}
Recording $H_e$ forms an initial bank $B_e^0$ before later goals are known. During execution, the same Writer appends observations without overwriting the past. At request time $t_q$, the Reader uses $B_e^q=\operatorname{Write}(H_e\Vert J_{e,<t_q})$ and selects $z_{e,q}=R_\phi(B_e^q,g_{e,q})$. Here $J_{e,<t_q}$ contains strict-past execution observations. The base policy $\theta_0$ is frozen, while retrieval parameters $\phi$ and action-interface parameters $\psi$ are learned.

\subsection{Memory Bank}\label{sec:archive}
The Writer comprises frozen visual encoding, feature sharing, temporal-entry creation and learned causal key computation. Each observation yields appearance features $A_i\in\mathbb R^{16\times1024}$. An appearance pool stores immutable prototypes $\bar A_u$ that multiple temporal entries can reference, avoiding repeated storage of matching features. Each entry keeps its own frame index, observed metadata, validity, prototype reference $p_i$, and retrieval key.

The Writer finds the nearest prototype by pooled cosine similarity and shares it only when both a similarity threshold and a maximum per-token distance bound are satisfied. Otherwise it appends a new prototype. The token-level check preserves locally distinctive observations that a pooled similarity could miss. This appearance sharing avoids duplicate feature storage while preserving separate temporal entries. Prototypes are never averaged or overwritten.

The original frame index and spatial grid supply the native positional term $T_i$. The reconstructed observation and its causal retrieval key are
\begin{equation}
 \widehat V_i=\bar A_{p_i}+T_i,\qquad
 k_i=E_\phi(\widehat V_{\leq i},\eta_{\leq i})\in\mathbb R^{256}.
 \label{eq:memory-bank}
\end{equation}
Here $\eta_i$ contains observed metadata, and $E_\phi$ uses only current and preceding entries. The key encoder is learned with the Reader; visual encoding and the sharing rule are fixed. Recording uses observations and metadata, not goal inputs or expert-action targets. Visual features preserve evidence for different later uses while matching the pretrained policy's visual interface. Appendix~\ref{app:memory-interface} gives the reconstruction, admission and representation details.

\subsection{Memory Retrieval}\label{sec:retrieval}
\begin{figure}[H]\centering\includegraphics[width=\textwidth]{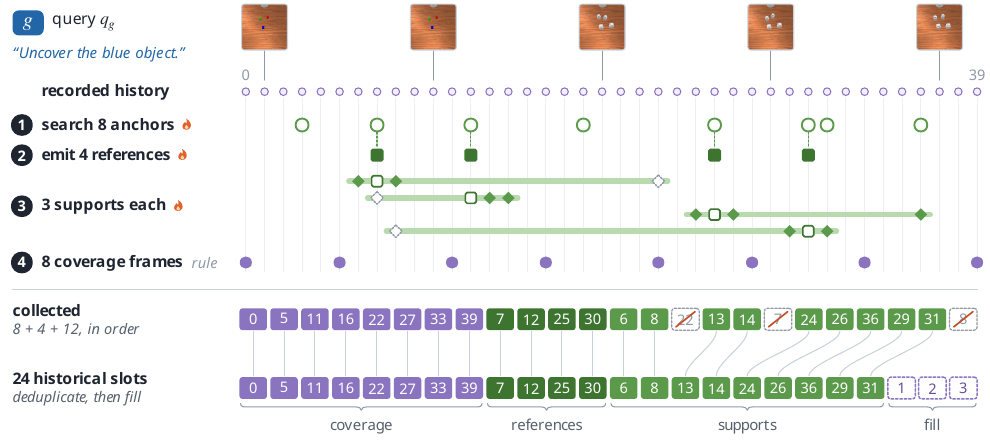}
\caption{\textbf{Retrieving complementary evidence.} The Reader searches candidate anchors, selects reference observations, and gathers temporally related support. Coverage frames preserve context across the request-time history. Deduplication and filling produce 24 historical slots for the action policy.}\label{fig:reader}
\end{figure}

A request can depend on several moments: remembered contents identify a case, while its earlier position supplies a return destination. The Reader retrieves complementary observations from the bank snapshot available when the request arrives (Figure~\ref{fig:reader}).

The instruction is pooled into a query over causal keys. The Reader searches eight candidate anchors, chooses four references, and selects three supporting observations for each reference using the goal and temporal relation. Eight evenly spaced coverage observations supplement these references and supports. Stable deduplication and temporal filling yield 24 historical slots. Histories shorter than this budget use masked padding.

A path $z$ records the discrete choices, and $I(z)$ gives the selected indices. Its score sums the masked log probabilities of the nonnull learned choices:
\begin{equation}
 S_\phi(z;B_e^q,g)=\sum_{j\in\operatorname{Dec}(z)}\log P_\phi(z_j\mid z_{<j},B_e^q,g).
 \label{eq:retrieval-packet}
\end{equation}
All requests retain the initial recording; later snapshots can also contain experience from preceding actions. Deployment takes the highest-scoring eligible choice at each decision. New writes become searchable at the next request, while the current request reuses its selected indices. Appendix~\ref{app:retrieval} gives the scoring functions, and Algorithm~\ref{alg:read} specifies the selection procedure.

\subsection{Action Generation}\label{sec:action}
\begin{figure}[H]\centering\includegraphics[width=\textwidth]{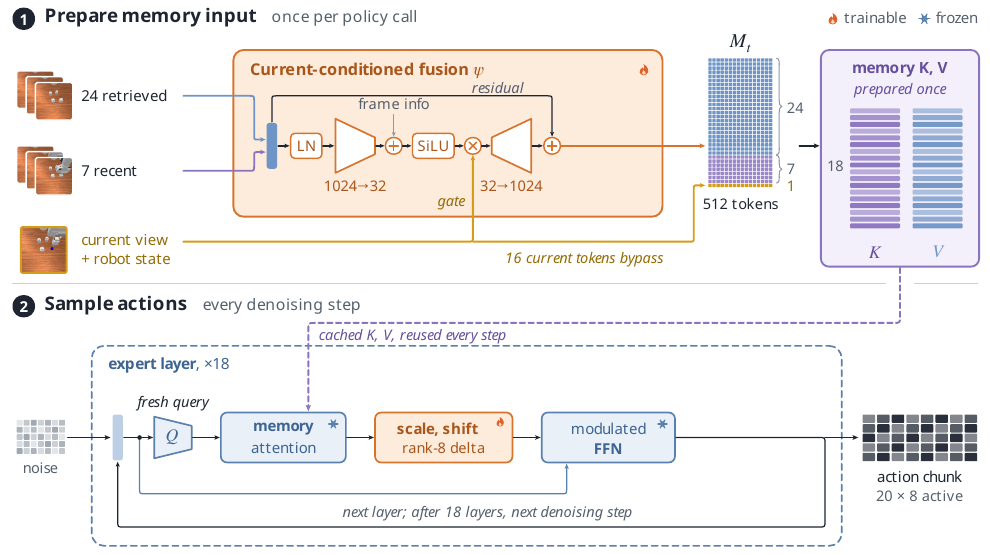}
\caption{\textbf{From memory to action.} Retrieved history and recent execution observations are fused with current context, then combined with current-frame tokens. The action expert generates fresh action queries using this memory input. Memory keys and values are prepared once per sampling call.}\label{fig:packet}
\end{figure}

Remembered identity and origin can remain useful after an object moves, but the next grasp must use its current location. The action input combines 24 retrieved observations, seven strict-past execution observations and the current observation. Recent context supplies immediate feedback even while the request's historical selection remains fixed. Learned fusion uses the current scene, robot state and recent context to adapt historical features for action (Figure~\ref{fig:packet}).

The memory input is $M_t=M_\psi(\widehat V_{I(z)},J_{e,<t},o_t,s_t)$. A width-32 residual fuses historical and recent features with current context. The current observation's native memory tokens bypass it. Each observation contributes 16 tokens, giving 512 tokens of width 1024.

The simulator policy is RoboMME's FrameSamp+Modul MME-VLA~\citep{robomme}, built on $\pi_{0.5}$~\citep{pi05}. Memory attention supplies scale and shift to the action expert. Fusion and rank-eight modulation deltas~\citep{lora} in 18 layers are adapted while native attention and the base policy remain frozen. SO101 uses the same memory architecture with a robot-specific prepared policy (Appendix~\ref{app:physical-protocol}).

Within a policy call, the memory input and validity mask are fixed. Each layer prepares memory keys and values once, while action queries and attention are recomputed at each denoising step $d$:
\begin{equation}
 (K_\ell,V_\ell)=\operatorname{Prepare}_{\theta_0,\ell}(M_t),\qquad
 c_{\ell,d}=\operatorname{Attn}_{\theta_0,\ell}
 (Q_{\theta_0,\ell}(x_{\ell,d}),K_\ell,V_\ell).
 \label{eq:memory-kv}
\end{equation}
Here $x_{\ell,d}$ denotes action features. The attention keys and values are projections of $M_t$, distinct from the bank's retrieval keys $k_i$. New observations or instructions rebuild these projections. Appendix~\ref{app:memory-interface} gives the fusion and modulation equations. Algorithm~\ref{alg:record-serve} summarizes the recording and execution procedure.

\subsection{Training}\label{sec:training}
\begin{figure}[H]\centering\includegraphics[width=\textwidth]{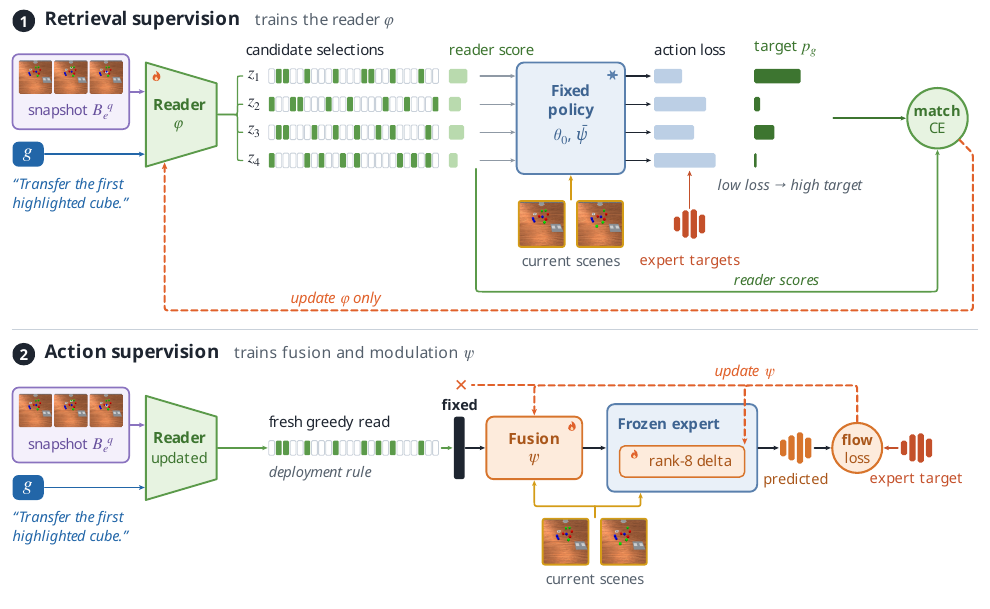}
\caption{\textbf{Learning retrieval and action use.} Expert-action prediction across current states sharing one request-time snapshot compares candidate evidence selections to supervise the Reader. The updated Reader then supplies the history used to train fusion and low-rank modulation, while the base policy stays frozen.}\label{fig:learning}
\end{figure}

Training teaches the Reader which past observations help predict expert actions, then teaches the action interface to use the Reader's selections. Each training group supplies a common request-time bank snapshot and several goals, with expert continuations from different current scenes (Figure~\ref{fig:learning}).

Within each group, $B_e^q$ is shared across its goals $\mathcal G_e$ and current-scene examples. Different execution prefixes form separate groups, so all candidate selections are legal for every example being compared. Goal $g$ supplies current-scene examples $m$ with positive weights $w_{g,m}$ summing to one. Let $\mathcal L_{g,m}(z;\psi)$ be the masked expert-action prediction loss for selection $z$. The historical facts remain the same across a goal's current-scene examples, while the required action changes.

For each goal, up to $C$ distinct selections are compared using a fixed action-interface snapshot $\bar\psi$. Expert targets, current inputs, noise and flow time are shared across candidate comparisons within an example. With $s_{g,c}=S_\phi(z_{g,c};B_e^q,g)$, the candidate losses give detached preferences and a retrieval loss:
\begin{equation}
 \begin{aligned}
 U_{g,c}&=\operatorname{sg}\!\left[\sum_mw_{g,m}\mathcal L_{g,m}(z_{g,c};\bar\psi)\right],\qquad
 p_g=\operatorname{sg}[\operatorname{softmax}(-U_g/T)],\\
 \mathcal L_\phi&=-\frac{1}{|\mathcal G_e^+|}\sum_{g\in\mathcal G_e^+}\sum_{c=1}^{C_g}
 p_{g,c}\log[\operatorname{softmax}(s_g)]_c.
 \end{aligned}
 \label{eq:utility-preference}
\end{equation}
Here $T>0$ is a temperature and $\operatorname{sg}$ stops gradients. The set $\mathcal G_e^+$ contains goals with informative preferences. No Reader update is made if it is empty. Goal-wise averaging balances the learning signal across different uses of the same experience.

After updating $\phi$, a fresh greedy read selects the evidence used to update the action interface:
\begin{equation}
 \begin{aligned}
 z_g^{\rm gr}&=R_\phi^{\rm greedy}(B_e^q,g),\\
 \mathcal L_\psi&=\frac{1}{|\mathcal G_e|}\sum_{g\in\mathcal G_e}\sum_mw_{g,m}
 \mathcal L_{g,m}(\operatorname{sg}(z_g^{\rm gr});\psi).
 \end{aligned}
 \label{eq:selected-action-update}
\end{equation}
This uses the same selection rule during action training and deployment. No action-loss gradient crosses the discrete selection, while the frozen policy transmits activation gradients to $\psi$. Algorithm~\ref{alg:learn} summarizes the two updates; Appendix~\ref{app:optimization} gives the action loss, candidates and initialization.

\begin{algorithm}[!htbp]
\caption{Learning retrieval and action use}\label{alg:learn}
\small
\textbf{Input:} Common request-time snapshot $B_e^q$, goals $\mathcal G_e$ and their weighted expert examples; reader $\phi$, action interface $\psi$; budget $C$, temperature $T>0$ and margin $\delta\geq0$.\par
\begin{tabularx}{\linewidth}{@{}rY@{}}
1 & $\bar\psi\gets\operatorname{sg}(\psi)$; $\mathcal G_e^+\gets\varnothing$.\\
2 & \textbf{for each} $g\in\mathcal G_e$ \textbf{do}\\
3 & \hspace*{.8em}$\mathcal Z_g\gets\operatorname{ProposeUnique}_\phi(B_e^q,g;C)$; $C_g\gets|\mathcal Z_g|$.\\
4 & \hspace*{.8em}Compute $U_{g,c}$ and $p_g$ by Eq.~\eqref{eq:utility-preference}, with fixed expert/noise operands.\\
5 & \hspace*{.8em}\textbf{if} $C_g>1$ and $(\max_cU_{g,c}-\min_cU_{g,c})/T>\delta$, add $g$ to $\mathcal G_e^+$.\\
6 & \textbf{end for}\\
7 & \textbf{if} $\mathcal G_e^+\neq\varnothing$: $\phi\gets\operatorname{OptStep}_\phi(\phi,\nabla_\phi\mathcal L_\phi)$ by Eq.~\eqref{eq:utility-preference}.\\
8 & Rebuild causal keys with updated $\phi$; $z_g^{\rm gr}\gets R_\phi^{\rm greedy}(B_e^q,g)$ for every $g$.\\
9 & $\psi\gets\operatorname{OptStep}_\psi(\psi,\nabla_\psi\mathcal L_\psi)$ by Eq.~\eqref{eq:selected-action-update}, with selections fixed.\\
10 & \textbf{return} $(\phi,\psi)$.\\
\end{tabularx}
\end{algorithm}

\FloatBarrier
\section{Experiments}\label{sec:experiments}
\subsection{Experimental Setup}\label{sec:setup}
\begin{figure}[!htbp]\centering
\includegraphics[width=\textwidth]{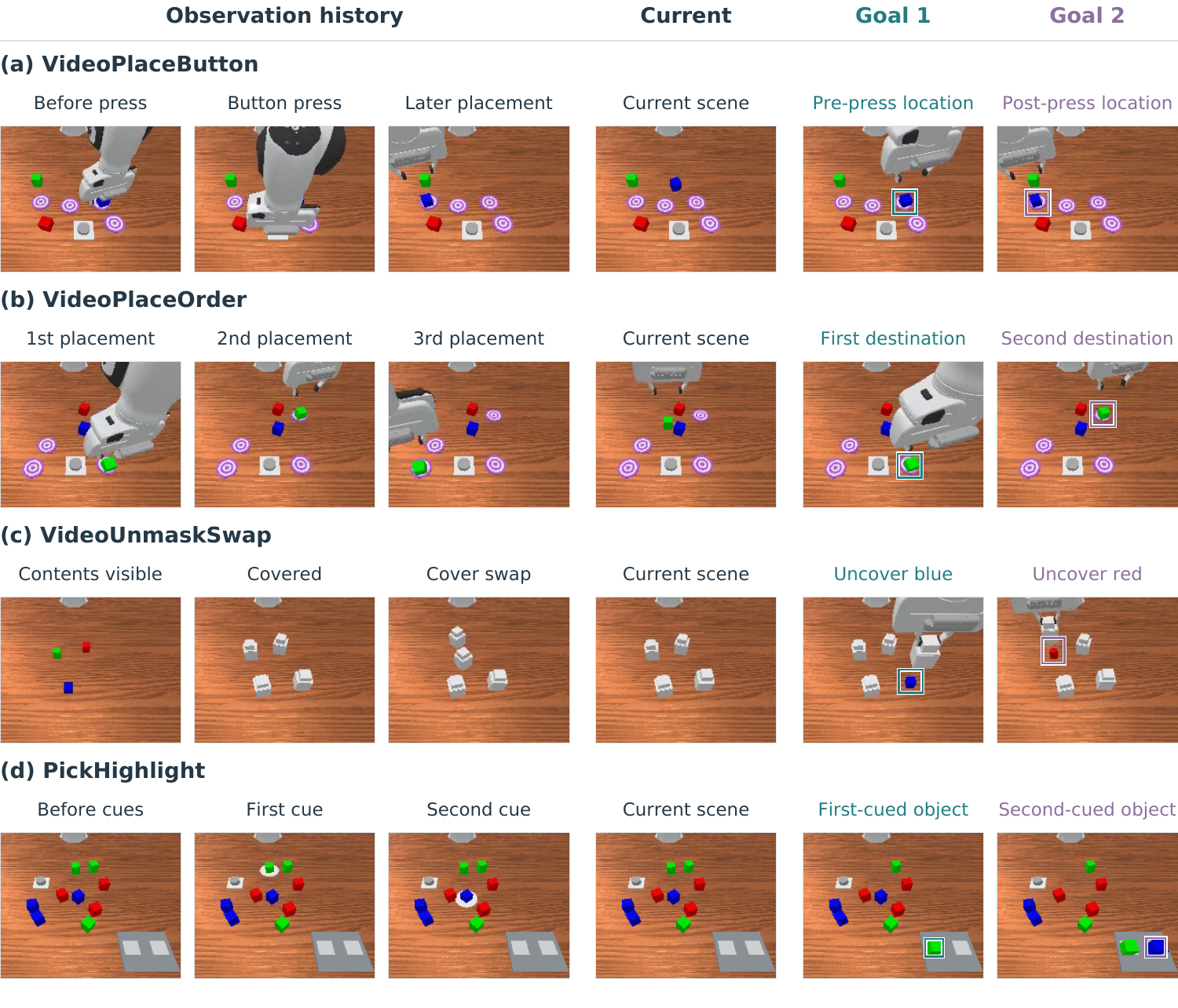}
\caption{\textbf{Simulation tasks and later goals.} Each row shows a common observation history, the current scene, and the manipulation specified by each later goal. The same past determines two destinations or targets through event timing, order, concealed identity or transient highlighting.}\label{fig:simulator}
\end{figure}

The experiments examine whether shared experience supports different later goals, whether reuse survives changes caused by earlier actions, and what storage and response costs it requires. Four RoboMME-based tasks~\citep{robomme} cover event-relative placement, temporal order, hidden object identity and transient visual cues (Figure~\ref{fig:simulator}). Each task uses 50 common recordings with two later goals, giving 400 separate goal executions and 200 separate two-request sequences per method. Two SO101 tabletop scenarios each use 20 recordings with three later requests.

Execution observations are appended through the Writer, and each new request reads the accumulated memory available at its arrival. Current and recent observations refresh the action input throughout execution.

\paragraph{Metrics.}
Goal success averages completion over individual requests. All-goal success counts a recording only if both requests succeed in independent branches restored to the same post-recording world and memory. Sequence success requires both requests without resetting the scene. All-goal isolates different uses of shared experience. Sequence additionally tests accumulated execution and changed current states. Physical evaluation reports the three independent requests, their joint All-goal success, and a separate continuous three-request sequence. Appendix~\ref{app:task-metrics} defines the metrics and task predicates.

\paragraph{Comparisons.}
No-Memory ($\pi_{0.5}$) receives current inputs. Uniform-Memory ($\pi_{0.5}$) selects 24 evenly spaced historical observations from its request-time bank snapshot with matched action adaptation. MME-VLA ($\pi_{0.5}$) uses released FrameSamp+Modul weights without retraining. MemER (RoboMME) uses the benchmark reproduction with remembered images and VLM-generated grounded subgoals. Internal No-Memory, Uniform-Memory and ReuseMem comparisons match the prepared policy, expert data and action-training budget. Released external systems retain their native memory and policy interfaces, so their differences reflect complete configurations. Physical comparisons share the same prepared SO101 interface. Appendix~\ref{app:comparison-ablations} gives the comparison configurations.

\paragraph{Training.}
Simulation uses 100 training, 20 validation and 50 test recordings per task. Physical scenarios use 50 training, 10 validation and 20 test demonstrations each. AdamW trains the Reader and action interface for up to 20,000 steps, with checkpoint selection based on validation performance. Appendix~\ref{app:training-settings} gives learning rates and batch composition.

\subsection{Simulation Results}\label{sec:simulator}
Table~\ref{tab:sim-main} reports individual, all-goal and sequential success. ReuseMem completes both independent uses of a recording in 121/200 cases (60.5\%), compared with 72/200 (36.0\%) for Uniform-Memory. It completes 116/200 continuous sequences (58.0\%), compared with 70/200 (35.0\%) for uniform selection. The improvements of 24.5 and 23.0 percentage points are obtained with matched memory construction and the same historical slot budget. They show the value of learning which past observations to provide to the adapted action interface.

\begin{table}[!htbp]\centering\small
\caption{\textbf{Success on four simulation tasks.} Each task contributes 50 common recordings with two later goals. Goal pools 400 independent requests; All-goal and Sequence each use 200 recordings. All-goal combines reset branches, whereas Sequence requires both requests without reset.}\label{tab:sim-main}
\begin{tabular}{lrrr}\toprule
Method & Goal (\%) & All-goal (\%) & Sequence (\%)\\\midrule
No-Memory ($\pi_{0.5}$) & 25.3 (101/400) & 5.5 (11/200) & 7.5 (15/200)\\
MME-VLA ($\pi_{0.5}$) & 32.5 (130/400) & 14.5 (29/200) & 11.5 (23/200)\\
MemER (RoboMME) & 39.3 (157/400) & 20.0 (40/200) & 20.5 (41/200)\\
Uniform-Memory ($\pi_{0.5}$) & 54.0 (216/400) & 36.0 (72/200) & 35.0 (70/200)\\
\rowcolor{pale}\textbf{ReuseMem} & 74.5 (298/400) & 60.5 (121/200) & 58.0 (116/200)\\
\bottomrule\end{tabular}
\end{table}

\begin{figure}[!htbp]\centering
\includegraphics[width=\textwidth]{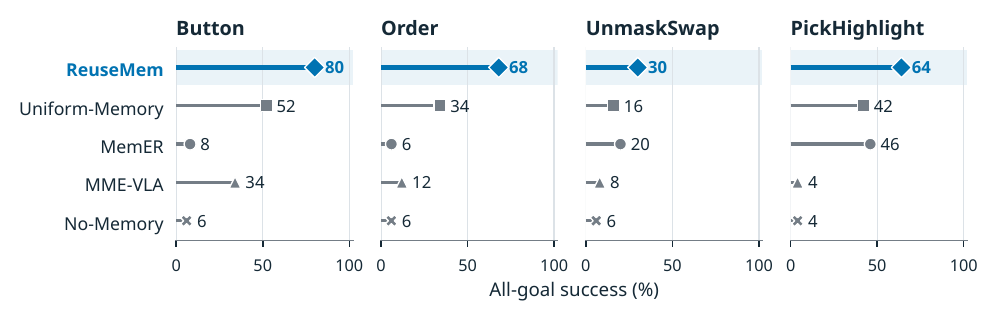}
\caption{\textbf{All-goal success by task.} Each point is the proportion of 50 common recordings for which both independent goals succeed. Learned evidence selection improves reuse across event-relative, temporal, hidden-identity and transient-cue tasks.}\label{fig:sim-results}
\end{figure}

Figure~\ref{fig:sim-results} shows improvements across all four types of historical cue. Hidden-object swaps remain the most challenging, with 30\% All-goal success for ReuseMem, compared with 64--80\% on the other tasks. This difficulty is consistent with the need to maintain target identity through occlusion and exchanges. All-goal tests different independent uses of a recording, while Sequence additionally tests action from states left by preceding requests. Appendix Table~\ref{tab:per-task} gives the individual counts.

\subsection{Ablation Studies}\label{sec:ablations}
\begin{figure}[!htbp]\centering
\includegraphics[width=\textwidth]{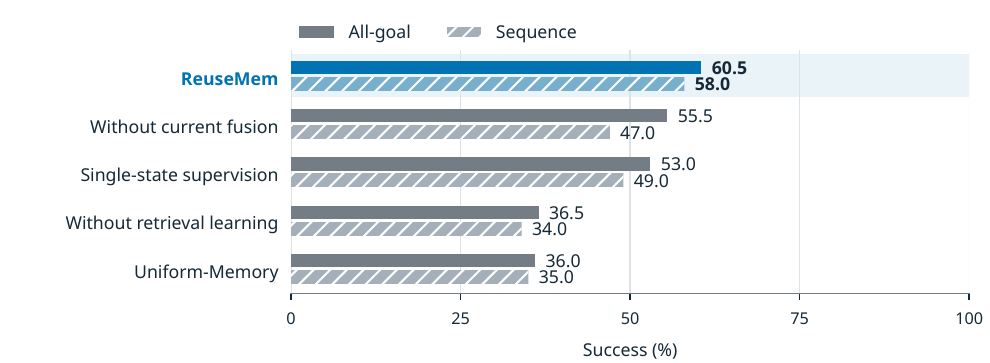}
\caption{\textbf{Learning retrieval and current-scene action use.} Bars compare independent All-goal success and no-reset Sequence success over 200 common recordings. Controls share the prepared policy, expert data and action-training budget.}\label{fig:ablation}
\end{figure}

The ablations separate learning access to the past from using it in the present (Figure~\ref{fig:ablation}). The variant without retrieval learning keeps the Reader and causal keys at their common initialization while retaining action adaptation. Single-state supervision forms retrieval preferences from one current state per goal while matching action-training data and budget. The variant without current fusion bypasses the historical/recent residual. Current images and modulation adaptation remain available.

Retrieval learning raises All-goal success from 36.5\% to 60.5\%. Using multiple current states for retrieval preferences improves on single-state supervision (53.0\%). Removing current fusion reduces All-goal success to 55.5\% and Sequence success to 47.0\%. Its larger effect on Sequence suggests that finding useful history and acting on it from a changed scene are distinct learning problems. Appendix~\ref{app:comparison-ablations} provides the complete ablation results. Appendix~\ref{app:evidence-use} checks whether selected observations cover the necessary facts and whether removing those facts disrupts action.

\subsection{Real-World Experiments}\label{sec:physical}
\begin{figure}[H]\centering
\makebox[\linewidth][c]{\includegraphics[width=1.045\linewidth]{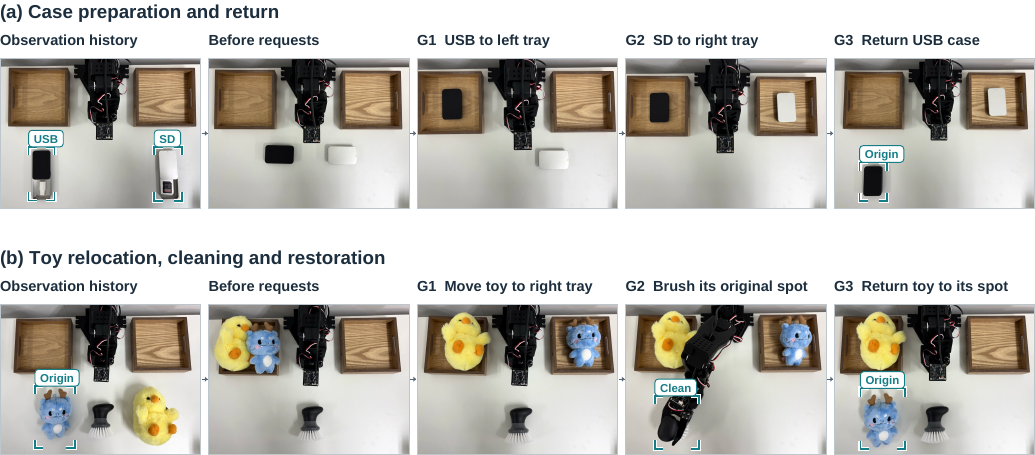}}
\caption{\textbf{Real-world manipulation sequences.} Each row shows a common observation history, the scene before requests, and three successive goals. (a) Case contents determine the transfer targets, and the recorded origin guides return. (b) The same toy origin supports relocation, localized brushing and restoration from the changing scene.}\label{fig:physical}
\end{figure}

Two tabletop scenarios use the same SO101 workspace and two trays (Figure~\ref{fig:physical}). Each common recording supports three requests. Independent branches test different uses of the history. Sequential execution preserves the previous actions\textquotesingle{} scene changes. Appendix~\ref{app:physical-protocol} gives the physical scoring and policy details.

\paragraph{Case preparation and return.}
The robot observes case contents and origins before the cases are closed and relocated. It transfers the USB case to the left tray and the SD-card case to the right tray, then returns the USB case to its recorded origin. Appearance, content assignment and original layout are counterbalanced.

\paragraph{Toy relocation, cleaning and restoration.}
The robot observes a blue dragon and a chick at desk positions before both are gathered in the left tray. It transfers the toy originally on the requested side to the right tray, brushes that toy\textquotesingle{}s original spot, parks the brush, and restores the toy. Original side and toy identity are counterbalanced.

\begin{table}[!htbp]\centering\small
\caption{\textbf{Physical manipulation success (\%).} Each scenario uses 20 common recordings. Goal averages 60 independent requests; All-goal and Sequence require all three independent or sequential requests, respectively.}\label{tab:physical}
\begin{tabular}{lrrrrrr}\toprule
Method & G1 & G2 & G3 & Goal & All-goal & Sequence\\\midrule
\multicolumn{7}{l}{\textit{Case preparation and return}}\\
No-Memory & 50 & 50 & 20 & 40.00 & 5 & 20\\
Uniform-Memory & 75 & 75 & 60 & 70.00 & 45 & 50\\
\rowcolor{pale}\textbf{ReuseMem} & \textbf{85} & \textbf{85} & \textbf{85} & \textbf{85.00} & \textbf{70} & \textbf{65}\\
\multicolumn{7}{l}{\textit{Toy relocation, cleaning and restoration}}\\
No-Memory & 50 & 45 & 45 & 46.67 & 10 & 10\\
Uniform-Memory & 80 & 75 & 75 & 76.67 & 60 & 55\\
\rowcolor{pale}\textbf{ReuseMem} & \textbf{90} & \textbf{85} & \textbf{85} & \textbf{86.67} & \textbf{70} & \textbf{65}\\
\bottomrule\end{tabular}
\end{table}

ReuseMem completes all three independent goals for 14/20 recordings in each physical scenario (Table~\ref{tab:physical}). It also finishes 13/20 continuous sequences in each, outperforming Uniform-Memory in both scenarios. These outcomes show the same recorded facts supporting object selection, restoration and tool use from changing current states. Visible-target controls achieve 19/20 successes for each of transfer, return and brushing, providing a reference for the underlying motor skills.

\subsection{Efficiency}\label{sec:storage}
\begin{figure}[H]\centering
\includegraphics[width=\textwidth]{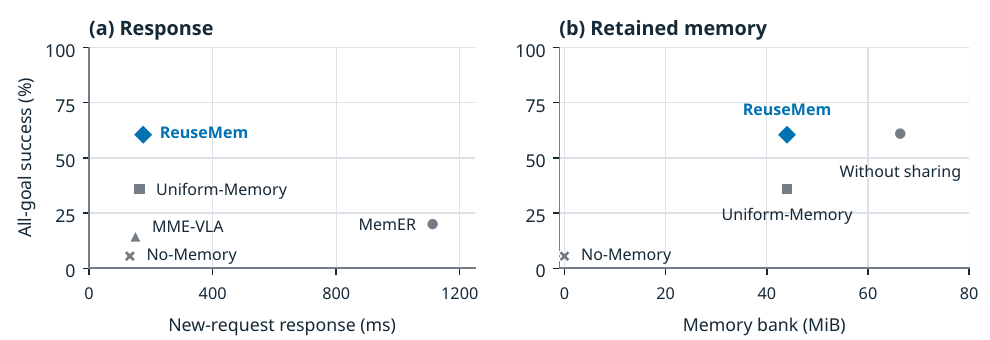}
\caption{\textbf{Success, response and retained memory.} Left: new-request response versus All-goal success. Right: bank storage for matched policy configurations at 2,048 observations. Sharing reduces stored visual content while retaining all temporal entries.}\label{fig:cost}
\end{figure}

Figure~\ref{fig:cost} relates All-goal success to new-request response and retained memory. ReuseMem responds in 175\,ms, compared with 163\,ms for Uniform-Memory and 1,112\,ms for MemER. Response includes the current input, memory processing, action generation and first-command dispatch. Measurements use an RTX 5090 with BF16 and batch size one. The complete ReuseMem path adds 12\,ms over uniform selection while improving All-goal success by 24.5 percentage points.

At a total of 2,048 temporal entries with 1,331 appearance prototypes, the retained bank occupies 43.98\,MiB rather than 66.38\,MiB with separate per-entry features, a 33.7\% reduction. All-goal success is 60.5\% with sharing and 61.0\% without it, a difference of one recording out of 200. The saving comes from storing matched appearances once, while retaining their separate temporal occurrences. Uniform-Memory uses the same bank representation and capacity, making its comparison a test of reading rather than storage capacity.

Within-call caching addresses a separate cost. For a continuing goal, memory projections reduce response from 174 to 165\,ms, a 9\,ms saving. New-request response, which includes a fresh retrieval step, is 175\,ms. Appendix~\ref{app:response-measurement} details response measurement and controls. Appendix~\ref{app:efficient} gives the bank and workspace accounting.

\FloatBarrier
\section{Conclusion}\label{sec:conclusion}
ReuseMem learns to use one visual memory for multiple later manipulation goals. A goal-independent Writer preserves experience, and learned retrieval and action fusion connect it to the current scene. Experiments show improved joint goal completion and continued reuse after earlier requests change the environment. Ablations and memory interventions link these gains to both selecting useful evidence and acting on it. Storing repeated visual features once and caching memory projections reduce the cost of reuse. Perceptual correspondence, precise manipulation and the bounded range of histories and goals remain limitations. Broader and longer interactions are a natural next step. This work connects the value of robot memory to the different actions that the same past experience can support.
\clearpage
{\fontsize{9}{10}\selectfont\setlength{\bibsep}{1.5pt}\bibliographystyle{unsrtnat}\bibliography{references/references}}
\clearpage\appendix
\section{Implementation Details}\label{app:implementation}
\subsection{Optimization}\label{app:training-settings}\label{app:optimization}
Table~\ref{tab:training-settings} gives the optimization settings. Each batch has four experience groups, with two associated goals and two current-scene examples per goal. Each example pairs the group's common request-time snapshot and revealed goal with current observations, robot state, recent execution context and an expert action chunk. The two examples for a goal share the snapshot and historical facts but differ in current inputs and expert continuations. Snapshots contain only observations preceding the request; future execution observations never enter candidate scoring. Retrieval and causal-key parameters $\phi$ use the Reader rate; fusion and modulation parameters $\psi$ use the action-interface rate. The model has 2,454,819 trainable Reader/index parameters and 747,944 action-interface parameters.

\begin{table}[!htbp]\centering\small
\caption{\textbf{Optimization settings.} Checkpoints are selected using validation performance; test recordings are reserved for final evaluation.}\label{tab:training-settings}
\begin{tabularx}{\textwidth}{lY}\toprule
Setting & Value\\\midrule
Optimizer & AdamW\\
Reader/index learning rate & $10^{-4}$\\
Action-interface learning rate & $5\times10^{-5}$\\
Batch size & 4 experience groups\\
Examples per group & 2 goals $\times$ 2 current states\\
Training budget & Up to 20,000 steps\\
Checkpoint selection & Validation performance\\\bottomrule
\end{tabularx}
\end{table}

\paragraph{Action loss.}
The simulator interface predicts 20-step chunks with 32 padded coordinates, of which eight are active. For example $q=(g,m)$, expert continuation $a_q^*$, valid-coordinate mask $\Omega_q$, Gaussian noise $\epsilon_q$ and flow time $\tau_q$, the loss is
\begin{align}
 a_q^{\tau_q}&=(1-\tau_q)a_q^*+\tau_q\epsilon_q,\qquad v_q^*=\epsilon_q-a_q^*,\notag\\
 \mathcal L_q(z;\psi)&=\frac{\sum_{h,j}\Omega_{qhj}(v_{qhj}-v^*_{qhj})^2}{\sum_{h,j}\Omega_{qhj}},
 \qquad\sum_{h,j}\Omega_{qhj}>0.
 \label{eq:reusemem-flow}
\end{align}
Here $v_q=v_{\theta_0,\psi}(a_q^{\tau_q},\tau_q\mid g,o_q,s_q,M_{\psi,q}(z))$ and $\mathcal L_q=\mathcal L_{g,m}$. Sampling integrates from $\tau=1$ to $\tau=0$. Each goal and current scene has its own expert continuation and legal recent observations. The physical policy uses its robot-specific action normalization and coordinates.

\paragraph{Candidate construction.}
Each goal uses up to $C=4$ candidates: the greedy path, an evenly spaced proposal, and seeded temporally stratified alternatives. Proposals obey the same masks as deployment. Evenly spaced choices traverse quantiles of the legal ordered pool. Stratified choices draw from successive temporal bins, falling back to the full pool when a bin is empty. Selected indices are removed within each selection stage. At most $8C$ proposals are attempted. Duplicate final ordered indices and validity masks are merged, retaining the first representative, so $1\leq C_g\leq C$. Prototype sharing alone does not merge candidates with different temporal occurrences.

A fixed snapshot $\bar\psi$ evaluates all candidates before either update, with dropout disabled and common expert targets, noise and flow times. A goal enters $\mathcal G_e^+$ when $C_g>1$ and $(\max_c U_{g,c}-\min_c U_{g,c})/T>\delta$. The temperature and margin are fixed before test evaluation, with the margin above fixed-input numerical variation. Reader scores are recomputed with gradients. The updated Reader then performs a fresh greedy read for the action loss. All goals contribute to the latter, including those without informative preferences. The optimizer skips empty or zero-gradient updates without changing its state.

\paragraph{Initialization and training cost.}
The fusion up-projection and second low-rank factor start at zero. Their paired down-projection and first factor are nonzero. This preserves the prepared policy while permitting learning. Temporal-coverage preparation uses the same expert data and counts toward the matched action-training budget. With two goals, two current states and four distinct candidates, one group requires 16 fixed-policy utility evaluations and four gradient-contributing action examples. These comparisons are training operations, not deployment work.

\subsection{Memory Representation and Action Interface}\label{app:memory-interface}
Table~\ref{tab:config} summarizes the memory and action-interface configuration.
\begin{table}[!htbp]\centering\small
\caption{\textbf{Model configuration.} Frame budgets, representation widths, and adaptation ranks have different roles.}\label{tab:config}
\begin{tabularx}{\textwidth}{YY}\toprule
Component & Setting\\\midrule
Appearance pool & Immutable $16\times1024$ BF16 appearance prototypes\\
Memory entry & Original ordinal, prototype reference, validity, metadata, and 256-dimensional key\\
Causal access index & Width 256; two blocks; four heads; 64-row local window\\
Reference/support search & Eight searched anchors; four references; three supports per reference\\
Historical packet & Eight coverage rows; stable deduplication/fill to 24 slots\\
Training group & Associated goals; two current-scene queries per goal\\
Training candidates & Up to four distinct historical selections per goal\\
Full memory packet & 24 historical + 7 strict-past execution + 1 current frame\\
Recording rate & 20 Hz\\
Action sampling & 10 denoising steps\\
Adaptation & Width-32 fusion; rank-8 deltas on 18 modulation kernels\\
Capacity & 2,048 retained rows across recording and execution\\\bottomrule
\end{tabularx}\end{table}

The key encoder projects each of the 16 spatial cells from width 1024 to 32, concatenates the ordered cells with observed metadata, and maps them to width 256. Two causal attention blocks use four heads and a 64-entry local window. The instruction query pools frozen language tokens through a learned projection, masked mean and attention pooling. Appearances use BF16 and retrieval keys FP32. The bank retains at most 2,048 entries with their original ordinals. Sharing appearances does not remove entries or increase this temporal capacity. Initial and execution observations share this budget. Archival admission stops when it is full, while recent context continues to update; no earlier entry is evicted or overwritten.

\paragraph{Appearance sharing.}
The frozen visual map separates appearance and positional terms:
\begin{equation}
 V_i=[P_i,H_i]W_{\rm out}+b_{\rm out}=A_i+T_i,\qquad
 A_i=P_iW_{\rm app},\quad T_i=H_iW_{\rm pos}+b_{\rm out}.
 \label{eq:shared-appearance}
\end{equation}
The positional term encodes the original frame ordinal and spatial grid. Both retrieval and action reconstruct observations from the same shared appearance and original position. For nonzero pooled appearances $\mu_i$ and $\bar\mu_u$, let $u^*$ maximize cosine similarity. Sharing is admitted only if
\begin{equation}
 \operatorname{cos}(\mu_i,\bar\mu_{u^*})\geq\tau_a,
 \qquad\max_{j=1}^{16}\|A_{ij}-\bar A_{u^*,j}\|_2\leq\delta_a.
 \label{eq:appearance-admission}
\end{equation}
Otherwise a new immutable prototype is stored. Thresholds are calibrated on validation recordings and fixed across test goals. Zero-norm pooled features share only an exact match. The per-token distance check limits substitution of local visual cues. Distinct temporal entries still retain separate keys, metadata and masks.

\paragraph{Fusion and modulation.}
Historical and recent tokens use the residual
\begin{equation}
 \widetilde v=v+W_{\uparrow}\!\left[\operatorname{SiLU}\!\left(W_{\downarrow}\operatorname{LN}(v)+h(u_i)\right)\odot\gamma(o_t,s_t,u_{\rm recent})\right].
 \label{eq:reusemem-fusion}
\end{equation}
Here $W_{\downarrow}:\mathbb R^{1024}\to\mathbb R^{32}$, $W_{\uparrow}:\mathbb R^{32}\to\mathbb R^{1024}$, and $h(u_i)$ encodes observed state, time and role. The sigmoid gate $\gamma$ uses current visual features, masked robot state and recent context. Invalid tokens remain masked, and current-observation tokens bypass fusion. At expert layer $\ell$, native memory attention supplies scale and shift:
\begin{align}
 c_\ell&=\operatorname{MemAttn}_{\theta_0,\ell}(x_\ell,M),&
 [\alpha_\ell,\beta_\ell]&=c_\ell(W_\ell^0+\Delta W_\ell)+b_\ell,
 \quad\Delta W_\ell=A_\ell B_\ell/r,\quad r=8,\label{eq:consumer}\\
 \widehat x_\ell&=\operatorname{RMS}(x_\ell)\odot(1+\alpha_\ell)+\beta_\ell.
 \label{eq:memoryrms}
\end{align}
The factors have shapes $A_\ell\in\mathbb R^{1024\times8}$ and $B_\ell\in\mathbb R^{8\times2048}$. Native attention, normalization and positional encoding remain fixed; only fusion and the low-rank factors are adapted. Memory keys and values are cached within one no-gradient sampling call. New inputs or parameter updates invalidate them. Action queries are recomputed at every denoising step, and training builds fresh computation graphs.

\subsection{Retrieval and Execution}\label{app:retrieval}\label{app:execution}
For retrieval key $k_i\in\mathbb R^{256}$ and pooled instruction $q_g$, anchor-search and reference-emission scores are
\begin{align}
 f^{\rm ref}_\phi(i;q_g,P)&=\frac{(W_rk_i)^\top[q_g+C_\phi(\operatorname{Mean}(\mathbf K_P))]}{\sqrt d},&
 f^{\rm emit}_\phi(i;q_g)&=\frac{(W_ek_i)^\top q_g}{\sqrt d},\quad d=256.\label{eq:reference-scores}
\end{align}
Here $C_\phi(x)=W_cx+b_c$ contributes zero when the previously searched set $P$ is empty. For reference $r$, support scores use
\begin{equation}
 \ell_{j\mid r,g}=\frac{(W_dk_j)^\top q_g+(W_ak_j)^\top B_\phi([q_g;k_r])}{\sqrt d}+w_\rho^\top\rho(j,r),\label{eq:support-scores}
\end{equation}
where $B_\phi(x)=W_bx+b_b$ and $\rho(j,r)$ encodes temporal difference and shared clock/role indicators. Anchors and emitted references exclude previously selected indices. Supports are distinct within each reference. Overlap across references is removed by final deduplication. Ties use the original frame ordinal. The path score is
\begin{equation}
 S_\phi(z)=\sum_{n\in\operatorname{Dec}(z)}\left[\ell_{\phi,n}(i_n)-\log\sum_{j\in\mathcal A_n}\exp\ell_{\phi,n}(j)\right].\label{eq:reusemem-path-score}
\end{equation}
Only learned choices contribute to this score. Algorithm~\ref{alg:read} gives the construction.
\begin{algorithm}[!htbp]
\caption{Reference and support retrieval}\label{alg:read}
\small
\textbf{Input:} Goal $g$, causal keys $\mathbf K$ from reconstructed entries, ordered legal indices $\mathcal I$, access weights $\phi$, choice rule $\mu$.\par
\noindent\textbf{Output:} Complete trace $z=(P,R,A,I_z)$ and neural score $S_\phi(z)$.\par\smallskip
\begin{tabularx}{\linewidth}{@{}rY@{}}
1 & $q_g\gets\operatorname{Pool}_\phi(g),\quad P\gets()$.\\
2 & \textbf{for} $u=1,\ldots,8$ \textbf{do}\\
3 & \hspace*{.8em}$i_u\gets\operatorname{Select}_\mu(f^{\rm ref}_\phi(\cdot;q_g,P),\mathcal I\setminus P);\quad P\gets P\Vert i_u$.\\
4 & \textbf{end for}\\
5 & $R\gets\operatorname{Select}^{(4)}_\mu(f^{\rm emit}_\phi(\cdot;q_g),P)$.\\
6 & \textbf{for} $u=1,\ldots,4$, with $r_u=R_u$ \textbf{do}\\
7 & \hspace*{.8em}$A_u\gets\operatorname{Select}^{(3)}_\mu(\ell_{\cdot\mid r_u,g},\mathcal I)$ if $r_u\neq\bot$; otherwise $A_u\gets(\bot,\bot,\bot)$.\\
8 & \textbf{end for}\\
9 & $I_z\gets\operatorname{Fill}_{24}\!\left(\operatorname{Cov}_8(\mathcal I)\Vert R\Vert\operatorname{Flatten}(A),\mathcal I\right)$.\\
10 & $z\gets(P,R,A,I_z),\quad s\gets S_\phi(z;\mathbf K,g)$ using Equation~\ref{eq:reusemem-path-score}.\\
11 & \textbf{return} $z,s$.\\
\end{tabularx}\par\smallskip
\noindent $I(z)=I_z$. $\operatorname{Select}^{(k)}$ makes $k$ sequential choices without replacement in its nonnull pool. An exhausted pool returns $\bot$ with zero score contribution. Serving uses greedy $\mu$; training proposals also use evenly spaced and temporally stratified choices. Masks and ordinal tie breaks are the same for proposal generation and neural rescoring.
\end{algorithm}

For legal ordered indices $\mathcal I=(i_0,\ldots,i_{N-1})$, coverage and filling are
\begin{align}
  n_b&=\min(b,N),\qquad
 \operatorname{Cov}_b(\mathcal I)=\left(i_{\lfloor u(N-1)/\max(1,n_b-1)\rfloor}\right)_{u=0}^{n_b-1},\label{eq:coverage}\\
 \operatorname{Fill}_b(L,\mathcal I)&=\operatorname{Pad}_b\!\left(\operatorname{Take}_b\!\left[\operatorname{Unique}^+\!\left(L\Vert\operatorname{Cov}_{\max(N,b)}(\mathcal I)\right)\right],\bot\right).\label{eq:fill}
\end{align}
The operation $\operatorname{Unique}^+$ removes nulls and repeated indices in first-occurrence order. Coverage precedes references and supports. Remaining legal observations fill unused slots before null padding. Null slots have zero features and false masks. The seven recent observations come only from execution before the current decision. They provide a separate fast-access context; the same observations can also be archived by the Writer for later requests. $J_e$ denotes the execution stream, not an additional retained full-frame archive.

Algorithm~\ref{alg:record-serve} uses the same Writer before and during execution. A snapshot is a read-only prefix of the bank at request receipt, not a duplicate of its appearance features. New entries preserve existing references and causal keys. Their eligibility begins with the next request snapshot, so historical indices can be reused within a request while the bank continues to grow. Every call refreshes current inputs, fusion and policy projections. Only newly observed evidence is archived; an issued command is not treated as an observed outcome. Goal transitions discard pending commands from the previous request.
\begin{algorithm}[!htbp]
\caption{Memory recording and policy execution}\label{alg:record-serve}
\small
\textbf{Input:} Recording $H_e$, subsequent requests and observations, parameters $(\theta_0,\phi,\psi)$, and flow grid $1=\tau_0>\cdots>\tau_D=0$.\par
\begin{tabularx}{\linewidth}{@{}rY@{}}
1 & Initialize an empty bank $B_e$, execution history $J_e\gets()$ and request index $q\gets0$.\\
2 & \textbf{for each} observation $(o_i,\eta_i)$ in $H_e$ \textbf{do}\\
3 & \hspace*{.8em}Compute frozen $A_i$; share or append its appearance by Eq.~\eqref{eq:appearance-admission}; append its temporal entry and causal key by Eq.~\eqref{eq:memory-bank}.\\
4 & \textbf{end for}; retain the initial prefix $B_e^0$.\\
5 & \textbf{for each} subsequent policy call $t$ \textbf{do}\\
6 & \hspace*{.8em}Acquire $(g_t,o_t,s_t)$; update recent context from newly available strict-past observations in $J_e$; record each once in $B_e$ using the Writer in line 3.\\
7 & \hspace*{.8em}\textbf{if} a new request arrives: discard pending commands; $q\gets q+1$; take the prefix $B_e^q$ strictly before request receipt and read $z_q\gets R_\phi^{\rm greedy}(B_e^q,g_t)$ by Algorithm~\ref{alg:read}.\\
8 & \hspace*{.8em}$M_t\gets M_\psi(\widehat V_{I(z_q)},J_{e,<t},o_t,s_t)$, including recent/current masks.\\
9 & \hspace*{.8em}Prepare $(K_\ell,V_\ell)$ once per layer by Eq.~\eqref{eq:memory-kv}; $a^{\tau_0}\sim\mathcal N(0,I)$.\\
10 & \hspace*{.8em}\textbf{for} $d=0,\ldots,D-1$ \textbf{do}\\
11 & \hspace*{1.6em}$a^{\tau_{d+1}}\gets a^{\tau_d}+(\tau_{d+1}-\tau_d)\,v_{\theta_0,\psi}(a^{\tau_d},\tau_d\mid g_t,o_t,s_t,M_t)$, reusing memory keys/values and recomputing action queries.\\
12 & \hspace*{.8em}\textbf{end for}; dispatch the scheduled commands from $a^0$.\\
13 & \textbf{end for}\\
\end{tabularx}
\end{algorithm}

\section{Evaluation Details}\label{app:protocol}\label{app:additional-evaluation}
\subsection{Tasks and Success Metrics}\label{app:task-metrics}
Each recording is committed before its later requests are supplied. Associated goals and current-state variants remain in the same data split. All methods receive camera observations, robot state, timestamps and the revealed request. Evaluator annotations such as object IDs, origins, cue intervals and success flags are withheld from the actor.

\paragraph{Simulation tasks.}
The four paired-goal variants reuse RoboMME scenes and event generators with request-specific completion predicates. In \textit{VideoPlaceButton}, the two destinations are the last placement before a button press and the first placement after it. \textit{VideoPlaceOrder} requests the first and second distinct recorded destinations. Destinations are free when requests begin, and execution does not change the recorded event order. In \textit{VideoUnmaskSwap}, the robot uncovers the requested blue or red object and releases its cover in a separate parking region. In \textit{PickHighlight}, two objects are highlighted at separate times before both cues disappear. Later requests transfer them to separate receiving slots. Each sequence leaves the next target accessible.

Let $Y_{e,g}$ be independent success for goal $g$ from recording $e$, and $Z_{e,k}$ success at sequential stage $k$. The metrics are
\begin{align}
 S_{\rm goal}&=\frac{\sum_e\sum_{g\in\mathcal G_e}Y_{e,g}}{\sum_e|\mathcal G_e|},&
 S_{\rm all}&=\frac1{|\mathcal E|}\sum_e\prod_{g\in\mathcal G_e}Y_{e,g},\label{eq:success}\\
 S_{\rm seq,K}&=\frac1{|\mathcal E|}\sum_e\prod_{k=1}^K Z_{e,k}.\label{eq:sequence-success}
\end{align}
Here $K=2$ in simulation and $K=3$ in the physical scenarios. All-goal is computed within each recording, not by multiplying averaged success rates. Independent branches restore the same post-recording world and initial bank. Sequences retain preceding scene changes and accumulated execution observations, with a new memory snapshot at each request. Each outcome is fixed when its request finishes. Failure or timeout ends a sequence, and unattempted later stages count as unsuccessful. A later return does not invalidate an earlier completed transfer.

\subsection{Physical Protocol}\label{app:physical-protocol}
The SO101 policy uses the same 18-layer memory-attention and modulation interface, initialized from MME-VLA. A robot-specific $\pi_{0.5}$ state/action interface and memory branch are prepared with SO101 expert continuations and temporal-coverage history, then frozen for Reader, fusion and rank-eight learning. Prepared weights, expert-data exposure and preparation budgets are matched across physical memory comparisons. Visible-target transfer, return and brushing trials assess the underlying motor skills.

Case contents and original positions are observed before the cases are closed and relocated. Transfer requires the correct closed case to be released in its assigned tray; return requires placement at its recorded origin. Origins are empty at request revelation. Appearance, content assignment and original layout are counterbalanced.

For the toy scenario, original side and toy identity are counterbalanced before both toys are gathered in a tray. The robot relocates the requested toy, brushes its emptied footprint, parks the brush and restores the toy. Localized brushing is scored by coverage of the recorded footprint and limited contact outside a calibrated margin. Sweeping both possible origins or the entire desk is rejected. Placement tolerances and execution deadlines are fixed using visible-target controls. Alternative origin acceptance regions are disjoint. These regions are used for scoring, not supplied to the policy.

\subsection{Comparisons and Ablations}\label{app:comparison-ablations}
Uniform-Memory uses $\operatorname{Cov}_{24}(\mathcal I)$ over its request-time snapshot, with the same Writer, appearance sharing, memory capacity, seven recent observations, current input, fusion and rank-eight adaptation as ReuseMem. No-Memory receives the current request, image and state without historical inputs. Released MME-VLA and the benchmark MemER reproduction retain their native memory and policy interfaces. Their scores use the same recordings and later requests, rather than published results from another benchmark.

The variant without retrieval learning keeps $\phi$ at the common initialization. Single-state supervision uses one current scene for retrieval preferences while retaining all action examples. The variant without current fusion bypasses the historical/recent residual while retaining current tokens and modulation adaptation. Data exposure, action-update counts and preparation budgets are matched. Tables~\ref{tab:per-task} and~\ref{tab:ablation-detail} report the per-task and ablation results.

\begin{table}[H]\centering\footnotesize
\caption{\textbf{Per-task simulation counts.} Every cell reports successes out of 50 common recordings. G1 and G2 use independent branches; All-goal requires both, and Sequence uses a separate two-request execution without reset.}\label{tab:per-task}
\begin{minipage}[t]{.485\textwidth}\centering
\begin{tabular}{@{}lrrrr@{}}\toprule
Method & G1 & G2 & All-goal & Sequence\\\midrule
\multicolumn{5}{l}{\textit{VideoPlaceButton}}\\
No-Memory & 13 & 14 & 3 & 4\\
MME-VLA & 29 & 27 & 17 & 13\\
MemER & 15 & 15 & 4 & 5\\
Uniform-Memory & 34 & 33 & 26 & 25\\
\rowcolor{pale}\textbf{ReuseMem} & 44 & 45 & 40 & 38\\
\midrule
\multicolumn{5}{l}{\textit{VideoPlaceOrder}}\\
No-Memory & 13 & 13 & 3 & 4\\
MME-VLA & 17 & 15 & 6 & 5\\
MemER & 13 & 13 & 3 & 4\\
Uniform-Memory & 30 & 24 & 17 & 18\\
\rowcolor{pale}\textbf{ReuseMem} & 42 & 40 & 34 & 34\\
\bottomrule\end{tabular}
\end{minipage}
\hfill
\begin{minipage}[t]{.485\textwidth}\centering
\begin{tabular}{@{}lrrrr@{}}\toprule
Method & G1 & G2 & All-goal & Sequence\\\midrule
\multicolumn{5}{l}{\textit{VideoUnmaskSwap}}\\
No-Memory & 13 & 13 & 3 & 4\\
MME-VLA & 12 & 11 & 4 & 3\\
MemER & 19 & 18 & 10 & 9\\
Uniform-Memory & 18 & 17 & 8 & 7\\
\rowcolor{pale}\textbf{ReuseMem} & 25 & 23 & 15 & 13\\
\midrule
\multicolumn{5}{l}{\textit{PickHighlight}}\\
No-Memory & 11 & 11 & 2 & 3\\
MME-VLA & 10 & 9 & 2 & 2\\
MemER & 32 & 32 & 23 & 23\\
Uniform-Memory & 31 & 29 & 21 & 20\\
\rowcolor{pale}\textbf{ReuseMem} & 40 & 39 & 32 & 31\\
\bottomrule\end{tabular}
\end{minipage}
\end{table}

\begin{table}[H]\centering\small
\caption{\textbf{Complete ablation results.} Goal pools 400 requests; All-goal and Sequence each use 200 recordings. Sufficient historical evidence supplies the annotated past evidence as a diagnostic control.}\label{tab:ablation-detail}
\begin{tabular}{lrrr}\toprule
Configuration & Goal (\%) & All-goal (\%) & Sequence (\%)\\\midrule
Without retrieval learning & 54.0 (216/400) & 36.5 (73/200) & 34.0 (68/200)\\
Single-state supervision & 69.0 (276/400) & 53.0 (106/200) & 49.0 (98/200)\\
Without current fusion & 70.5 (282/400) & 55.5 (111/200) & 47.0 (94/200)\\
\rowcolor{pale}\textbf{ReuseMem} & 74.5 (298/400) & 60.5 (121/200) & 58.0 (116/200)\\
Without feature sharing & 74.8 (299/400) & 61.0 (122/200) & 58.0 (116/200)\\
Without within-call KV cache & 74.5 (298/400) & 60.5 (121/200) & 58.0 (116/200)\\
Sufficient historical evidence & 82.5 (330/400) & 72.0 (144/200) & 68.5 (137/200)\\
\bottomrule\end{tabular}
\end{table}

\subsection{Historical Evidence and Action Use}\label{app:evidence-use}
Evidence coverage requires the selected 24 observations to contain every necessary annotated fact group. Equivalent views are interchangeable. Table~\ref{tab:evidence-selection} separates this retrieval diagnostic from manipulation success.

\begin{table}[!htbp]\centering\small
\caption{\textbf{Evidence selection and action success.} Coverage requires the selected 24 historical observations to cover every necessary annotated fact group; equivalent views within a group are interchangeable. Each metric uses 400 goal queries.}\label{tab:evidence-selection}
\begin{tabular}{lrr}\toprule
Method & Evidence coverage (\%) & Goal success (\%)\\\midrule
Uniform-Memory & 80.5 (322/400) & 54.0 (216/400)\\
Without retrieval learning & 75.0 (300/400) & 54.0 (216/400)\\
Single-state supervision & 87.5 (350/400) & 69.0 (276/400)\\
ReuseMem & 90.0 (360/400) & 74.5 (298/400)\\
\bottomrule\end{tabular}
\end{table}

\begin{figure}[H]\centering
\includegraphics[width=\textwidth]{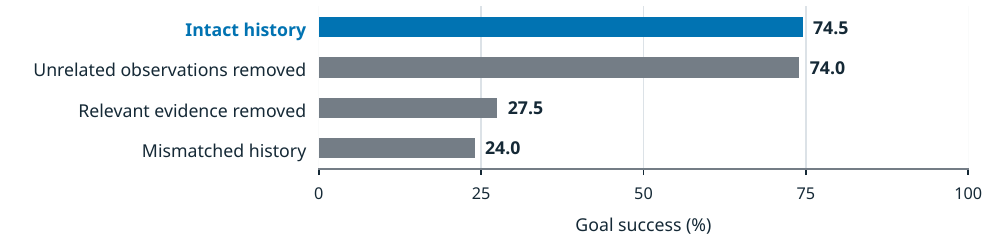}
\caption{\textbf{Dependence on historical evidence.} Removing relevant observations or supplying mismatched history reduces goal success more than removing unrelated observations. All conditions retain the original task truth and current inputs.}\label{fig:evidence-use}
\end{figure}

Memory interventions hold the current scene, request and task truth fixed while removing relevant or unrelated observations, or supplying a mismatched recording. Historical slot count and recent/current inputs are unchanged. Figure~\ref{fig:evidence-use} shows that removing relevant evidence or replacing history has a much larger effect than removing unrelated observations.

\section{Runtime and Memory Details}\label{app:timing}
\subsection{Response Measurement}\label{app:response-measurement}
Response time runs from receipt of a request to the first dispatched command computed for it. It includes input preparation, retrieval, reconstruction, fusion, action generation and dispatch. Measurements use one RTX 5090 GPU, BF16 computation and batch size one, averaged over 200 calls after 20 warm-up calls. Models and memory are resident. Loading, compilation, recording and physical motion are outside this interval. Table~\ref{tab:timing-stages} gives the measured stages.

\begin{table}[!htbp]\centering\footnotesize
\renewcommand{\arraystretch}{0.90}
\caption{\textbf{New-request response components.} Components sum to each method\textquotesingle{}s complete instruction-to-first-command latency.}\label{tab:timing-stages}
\begin{tabular}{lr}\toprule
Stage & Time (ms)\\\midrule
\multicolumn{2}{l}{\textit{ReuseMem}}\\
Observation, I/O and dispatch & 90\\
Current vision and language & 37\\
Reader & 11\\
Reconstruction, fusion and memory K/V & 10\\
Action sampling & 27\\
\textbf{Total} & \textbf{175}\\\addlinespace
\multicolumn{2}{l}{\textit{MemER (RoboMME)}}\\
Observation and image transfer & 33\\
Frame assembly and preprocessing & 90\\
VLM vision and prefill & 258\\
VLM decoding & 600\\
Parsing and memory update & 3\\
Low-level input preparation & 43\\
Low-level policy inference & 69\\
Command dispatch & 16\\
\textbf{Total} & \textbf{1112}\\\addlinespace
\bottomrule\end{tabular}
\end{table}

\begin{figure}[H]\centering
\includegraphics[width=\textwidth]{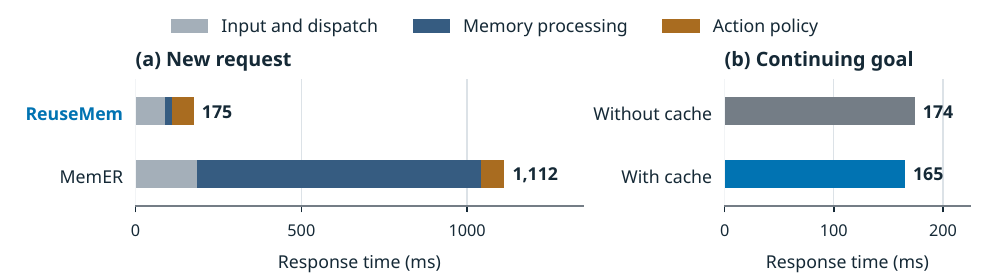}
\caption{\textbf{Response composition and within-call caching.} Left: instruction-to-first-command response split into input/dispatch, memory processing and action-policy computation. Memory processing includes the high-level VLM in MemER. Right: memory-K/V caching compares continuing-goal calls and reduces mean response by 9\,ms.}\label{fig:latency-details}
\end{figure}

Within a request, historical indices from its snapshot are reused while current features, fusion and policy projections are refreshed. The within-call cache control compares this same continuing-goal operating point (Figure~\ref{fig:latency-details}), reducing response from 174 to 165\,ms. It is separate from the 175\,ms response to a new request.

Without appearance sharing, new-request response is 172\,ms. Recording takes 10\,ms per observation including visual encoding, or 2\,ms once visual features are available.

\subsection{Memory Accounting}\label{app:efficient}\label{app:efficient-memory}
For $N$ entries, $U$ prototypes, $F=16\cdot1024$ appearance elements, retrieval width $d=256$, value/key element sizes $s_v,s_k$, per-entry metadata $m$ and prototype-reference bytes $p$, bank storage is
\begin{equation}
 B_{\rm shared}=UFs_v+N(ds_k+m+p)+B_{\rm lookup}+B_{\rm writer},\qquad
 B_{\rm dense}=N(Fs_v+ds_k+m)+B_{\rm writer}.
 \label{eq:shared-storage}
\end{equation}
Here $B_{\rm writer}$ is streaming Writer K/V and $B_{\rm lookup}$ is persistent appearance-lookup storage. The default bank recomputes pooled lookup descriptors, so $B_{\rm lookup}=0$. Shared appearances replace per-entry values rather than being retained alongside them.

With 2,048 entries and 1,331 prototypes, bank storage is 43.98\,MiB, compared with 66.38\,MiB without sharing. One MiB is $2^{20}$ bytes. Policy input and action-time memory K/V need 2 and 9\,MiB per call. These workspaces, model weights and allocator overhead are excluded from retained-bank storage.

\end{document}